\documentclass[manuscript]{acmart}
\usepackage{colortbl}
\usepackage{subcaption}
\usepackage{float}
\usepackage{multirow}
\usepackage{soul}
\usepackage{wrapfig}

\AtBeginDocument{%
  \providecommand\BibTeX{{%
    \normalfont B\kern-0.5em{\scshape i\kern-0.25em b}\kern-0.8em\TeX}}}

\setcopyright{none}
\acmDOI{10.1145/3654705}

\newcommand{\kb}[1]{\color{black} #1 \color{black}}

\begin{document}

\title{Using Deep Learning to Increase Eye-Tracking Robustness, Accuracy, and Precision in Virtual Reality}

\author{Kevin Barkevich}
\email{kdb2713@rit.edu}
\orcid{0009-0006-6174-8348}
\affiliation{%
  \institution{Rochester Institute of Technology}
  \country{USA}
}

\author{Reynold Bailey}
\email{rjbvcs@rit.edu}
\orcid{0000-0001-8964-9663}
\affiliation{%
  \institution{Rochester Institute of Technology}
  \country{USA}
}

\author{Gabriel J. Diaz}
\email{gjdpci@rit.edu}
\orcid{0000-0002-1812-017X}
\affiliation{%
  \institution{Rochester Institute of Technology}
  \country{USA}
}


\begin{abstract}
Algorithms for the estimation of gaze direction from mobile and video-based eye trackers typically involve tracking a feature of the eye that moves through the eye camera image in a way that covaries with the shifting gaze direction, such as the center or boundaries of the pupil. Tracking these features using traditional computer vision techniques can be difficult due to partial occlusion and environmental reflections. Although recent efforts to use machine learning (ML) for pupil tracking have demonstrated superior results when evaluated using standard measures of segmentation performance, little is known of how these networks may affect the quality of the final gaze estimate. This work provides an objective assessment of the impact of several contemporary ML-based methods for eye feature tracking when the subsequent gaze estimate is produced using either \kb{feature}-based or model-based methods. Metrics include the accuracy and precision of the gaze estimate, as well as drop-out rate.
\end{abstract}




\maketitle

\section{Introduction}

Although many researchers that use eye tracking are motivated to explore gaze behavior outside of the laboratory, in more natural experimental contexts, some will hesitate when faced by the large differences in data quality between desktop (i.e. remote) and head-mounted (i.e., mobile) eye trackers. To a large degree, the difference in quality between remote and head-mounted eye trackers is related to the ability for each eye tracker to accurately identify features in the eye image, such as the iris \cite{Chaudhary2019b} or pupil boundary or centroid \cite{Swirski2012, Kassner2014, Fuhl2015a, Fuhl2015b, Javadi2015, Santini2017, santini2018purest}  --- features that are informative because they move through the eye image in a way that covaries with the shifting gaze direction. 
The tendency for mobile eye trackers to use smaller eye cameras and to capture lower resolution eye images for the purpose of reducing size, power consumption, and latency has consequences on the subsequent gaze estimate. The problem is exacerbated by the need to place the eye cameras at far oblique angles in order to minimize the occlusion of the wearer's field of view \cite{Fuhl2016}. Unlike remote eye trackers that operate under controlled lighting conditions, mobile eye trackers often suffer from data dropouts when the infrared eye cameras are unable to sufficiently compensate for the dynamic range of ambient infrared illumination in the natural environment. 

\begin{wrapfigure}{R}{0.6\textwidth}
    \vspace{-4mm}
    \includegraphics[width=\linewidth]{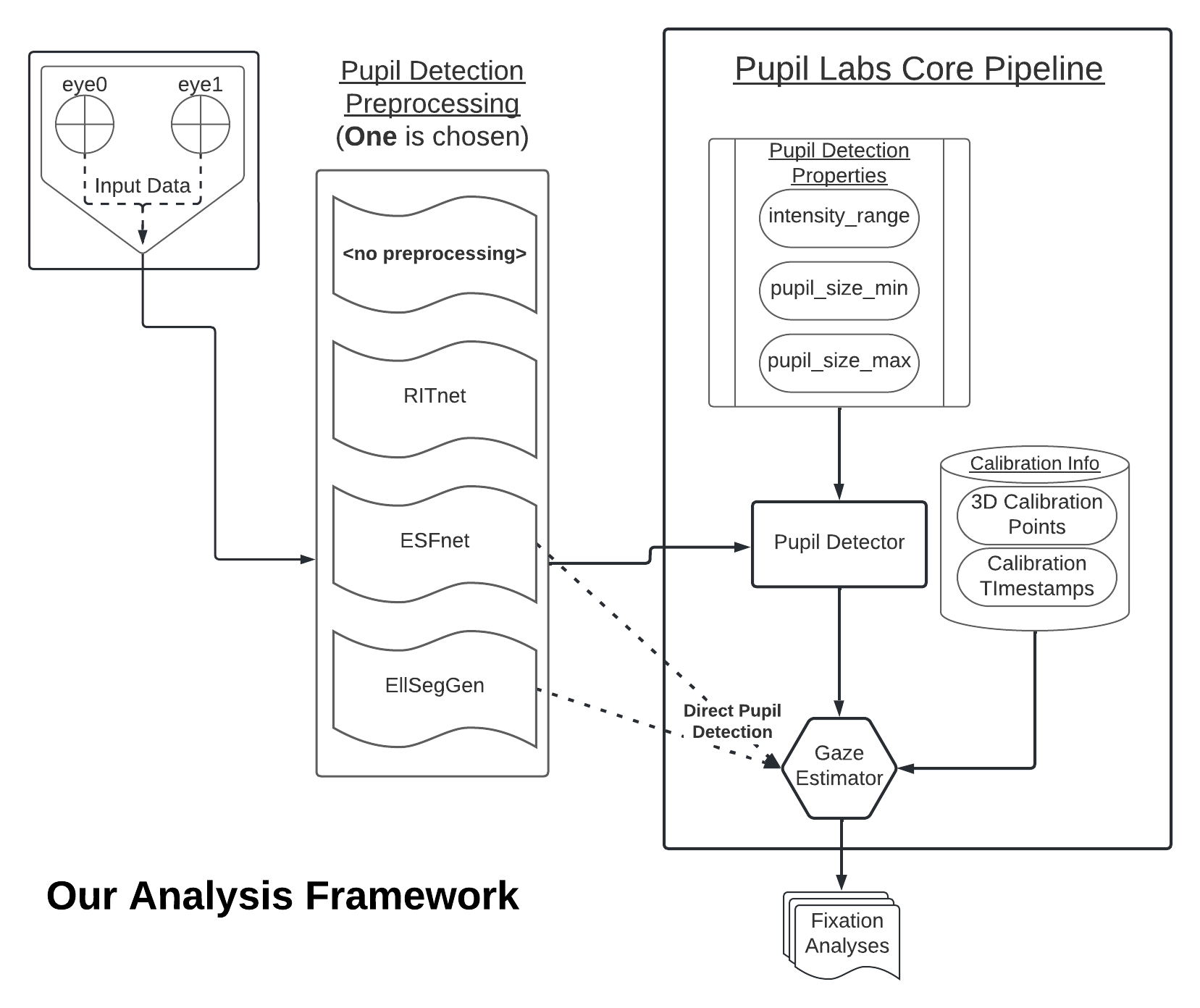}
    \caption{The pipeline through which our experiment data is processed, starting from the input eye images/video frames and feeding into our analysis of the gaze estimations. By default the eye images are not preprocessed by a neural network. We aim to explore the impact of using various neural network-based feature detection techniques in the preprocessing phase. We also aim to explore the impact of several of these neural networks when used to directly output pupil locations, bypassing the Pupil Labs pupil detector.}
    \label{fig:pl-pipeline-flowchart}
\end{wrapfigure}

Given the importance of feature tracking in the algorithmic process of video-based gaze estimation, it may be unsurprising that many laboratories have been exploring the use of machine learning for accurate eye image segmentation and feature localization \cite{Chaudhary2019, Kothari2020, Kothari2022, Cai2021, Fuhl2021, Wang2021}. What \textit{is} surprising is that progress in this area has had minimal impact on the accuracy of consumer level mobile eye tracking systems, or on public interest or adoption rates of mobile eye tracking. In part, this is because little has been done to measure the effect that improvements to the accuracy of the feature localization stage will have upon the quality of the final gaze estimate. This study aims to \kb{introduce and use a custom pipeline (Fig. \mbox{\ref{fig:pl-pipeline-flowchart}}) to} provide an objective evaluation of the contribution of improved feature detection models on the quality of the final gaze estimate \kb{when applied to a widely adopted open-source eye tracking solution. For the purposes of this study, this gaze estimate is obtained through the use of the open-source Pupil Labs gaze mapping software \mbox{\cite{Pupil}}. Two algorithms were used for gaze mapping. The Pupil Labs feature-based gaze mapper was included as it provides a simple polynomial mapping between pupil location and gaze direction, and because its straightforward and transparent nature suggests that results will generalize well to other gaze mapping systems. In addition, we tested with a more contemporary but opaque algorithm developed to address some of the known shortcomings of the simple polynomial mapping: the Pupil Labs 3D model-based gaze mapper.} Our general approach involves testing and reporting on the impact of several contemporary eye segmentation networks on the spatial accuracy, precision, and robustness to dropouts of the final gaze estimate, while other properties of the eye tracking pipeline remain unchanged.

\section{Background}
\subsection{Sources of error in pupil detection}

There are many sources of inaccuracy and imprecision that can impact pupil detection. These can stem from hardware limitations such as low spatial/temporal camera resolutions and extreme off-axial camera angles \cite{Fuhl2016} (which is especially prevalent inside virtual reality headsets), or from environmental factors such as reflections on the surface of the eye or poor lighting. Individual differences such as the amount of contrast between the pupil and the iris when imaged in the near-infrared, and obstructions from eyelashes and eyelids also play a role \cite{Fuhl2016}. Some algorithms for pupil segmentation, including the algorithm adopted by Pupil Labs  \cite{Kassner2014, Pupil2}, involve segmentation of the gray-scale histogram and rely on the assumption that the pupil is the darkest set of pixels within a region of interest in the eye-image. It is for this reason that eyelashes can be problematic, especially when darkened and made thicker by mascara. Some of these issues, such as low spatial/temporal camera resolutions and poor camera angles, cannot be easily overcome using only software. Many issues, however, may be compensated for with better algorithms that are more robust to occlusion, low contrast, and varying lighting conditions.

\subsection{The effect of feature localization accuracy on the gaze estimate}

The specific effect that degraded feature detection may have upon the subsequent gaze estimate will differ depending on the nature of the subsequent gaze estimation algorithm. Many video based eye trackers adopt \textit{\kb{feature}-based} algorithms \cite{mackworth1962head, merchant1967oculometer} which model the direct relationship between the movement of features in the eye image (e.g. the pupil centroid) and the location of gaze on an outward facing scene camera. Although \kb{feature}-based algorithms were introduced in the 1960's, some contemporary eye trackers, including the Eyelink 1000 \cite{SRResearch} still rely upon \kb{feature}-based methods. Because \kb{feature}-based algorithms do not typically condition the immediate estimate on the basis of prior information (with the occasional exception, such as \cite{Li2005}), an inaccurate feature localization will have an immediate and direct influence on accuracy through an inaccurate mapping to the gaze location within scene camera coordinates. However, some newer systems instead utilize \textit{\kb{3D} model-based} algorithms for gaze estimation, which monitor the movement of  features for the purpose of accumulating evidence that, through a process of error-minimization, allows them to refine the estimated pose of a 3D geometric model of the human eye within eye-camera space \cite{swirski2013fully, Kassner2014}. The contribution of poor feature localization on the gaze estimate of a \kb{3D} model-based system is less direct, because features serve two purposes:  1) at a relatively low temporal frequency, eye features contribute to the incremental updating of the model used for subsequent gaze estimates, and 2) at a higher temporal frequency, the eye feature will then be projected upon this 3D geometric eye-model for the purpose of ray-casting (e.g. gaze from the eyeball center through the pupil center). \kb{This information about temporal update rate is important when attempting to diagnose the cause of an inaccurate gaze estimate. If a gaze estimate is inaccurate on a specific frame, it is difficult to attribute the cause to an inaccurate pupil segmentation on that frame, or to a 3D eye model that was poorly fit to unreliable data from preceding frames.}

\subsection{Machine learning methods for eye image segmentation}
\begin{figure}
    \includegraphics[width=\linewidth]{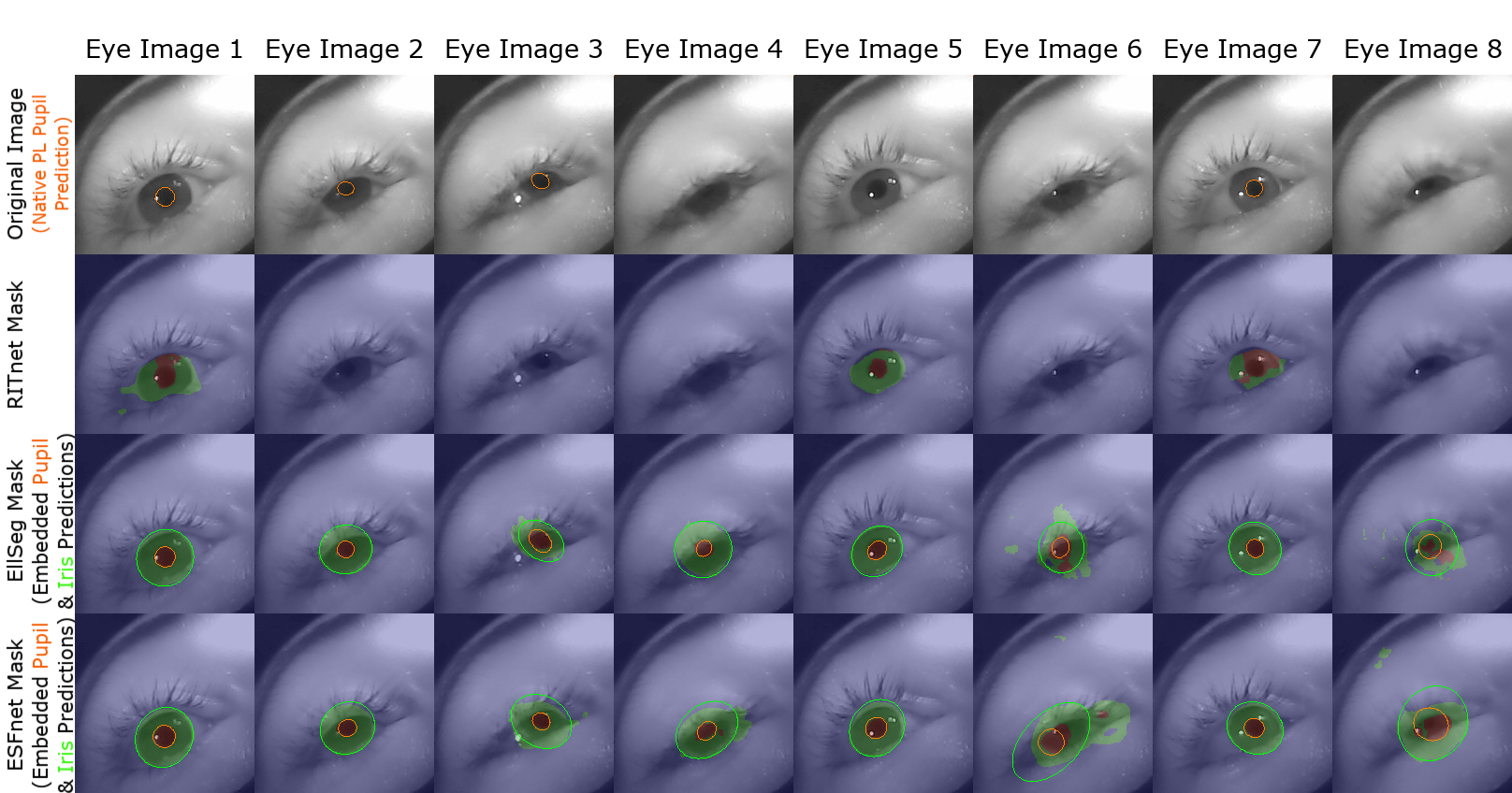}
    \caption{Comparison of different semantic segmentation techniques applied to 192x192px eye images captured during our VR data collection sessions. \textbf{Top:} original images with Pupil Labs\kb{'} default pupil prediction (orange). \textbf{Rows 2, 3, 4:} semantic segmentation results for neural network-based techniques \textit{RITnet}, \textit{EllSegGen}, and \textit{ESFnet} respectively. \textit{EllSegGen} and \textit{ESFnet} are also capable of directly predicting the ellipse parameters that encapsulate the pupil (orange) and iris (light green).}
    \label{fig:ritnet-ellseg-mask}
\end{figure}

Eye feature segmenting solutions attempt to provide a semantic label for each pixel in the image corresponding to the various parts of the eye (e.g. pupil, iris, sclera, skin/other). Machine learning-based approaches to this type of semantic segmentation have progressed significantly in recent years, with the availability of eye feature semantic segmentation datasets \cite{Garbin2019} giving rise to neural networks designed to locate the pupil on a per-pixel level \cite{Yiu2019, Chaudhary2019, Kothari2020, Kothari2022, Wang2021}. Two state-of-the-art systems include \textit{RITnet} \cite{Chaudhary2019} and \textit{EllSeg}/\textit{EllSegGen} \cite{Kothari2020, Kothari2022}. The outputs of \textit{RITnet} and \textit{EllSegGen} differ from each other in that while \textit{RITnet} always predicts pixel-for-pixel masks of what category of eye feature is shown in the input image, \textit{EllSegGen} (as well as \textit{ESFnet} \cite{Wang2021}, a similarly structured neural network) predicts the location of its eye features even when they are obscured by obstructions such as eyelashes and the upper eyelid (Fig. \ref{fig:ritnet-ellseg-mask}). Despite these semantic segmentation solutions being high-performing, they have not yet been widely adopted for use in the field of eye-tracking.

There are several possible explanations for the lag between progress in feature detection and its incorporation into commercial or open-source eye tracking pipelines. \kb{One possibility is that those who might be interested in the technology, including the broader scientific community, do not have the technical foundation to understand the connection between feature extraction and downstream real-world eye tracking performance.} A common metric for measuring the quality of a semantic segmentation neural network is the intersection over union (IoU) score, which is a ratio of the number of pixels that were predicted correctly over the number of pixels with that label overall. The closer this score is to 100 (since this ratio is commonly multiplied by 100), the better the neural network is said to have performed. Although the modern pupil segmenting solutions that have been discussed are demonstrated to be very effective at locating the pupil in a broad range of eye images, with mean IoU (mIoU) scores across semantic labels (e.g., pupil, iris, sclera, other) as high as 95.3 \cite{Chaudhary2019}, the potential impact that these improved feature tracking algorithms have on the quality of the final gaze estimate has not been well characterized. It is also notable that many of these networks increase the computational load, and not all are able to run in real-time. 

Intuitively, since semantic segmentation neural networks are capable of filtering out unwanted information such as poor lighting and the occlusion of eye features, and were trained on eye images representing a large variety of different eye shapes and colors, we expect that their use as a pre-processing step will improve the accuracy, precision, and dropout rate of the final gaze estimate. Below, we present the results of an objective evaluation of the contribution of improved feature detection models on the quality of the final gaze estimate in a popular open-source eye tracker integration into virtual reality by Pupil Labs. The study makes the following contributions:
\begin{itemize}
    \item We provide an open-source pipeline for the batch processing of eye tracking videos to obtain gaze estimates using one or more segmentation neural networks as a feature detector.
    \item Using this pipeline, we provide an objective evaluation of three neural networks (\textit{RITnet}, \textit{EllSegGen}, and \textit{ESFnet}) when compared to the widely-used, open-source, commercial feature detector provided by Pupil Labs.
    \item We provide concrete, data-driven recommendations on which feature detection neural network to use in terms of dropout rate, accuracy, and precision.
\end{itemize}

\section{Methodology}

\subsection{Privacy \& Ethics Statement}

In this IRB-approved study, subjects gave informed consent before participating. They were given the option to end data collection at any time. During the setup, the VR headset and eye-tracker hardware were described to the participant and they were allowed to ask questions before data collection began. 

\subsection{Participants}
\label{subsec:participants}

10 participants (2 females, 7 males, and 1 that did not indicate) volunteered to participate in this study. All participants reported normal and corrected-to-normal vision with no color vision abnormalities.
\subsection{Hardware \& Data Collection}

\begin{wrapfigure}{r}{0.45\textwidth}
    \includegraphics[width=\linewidth]{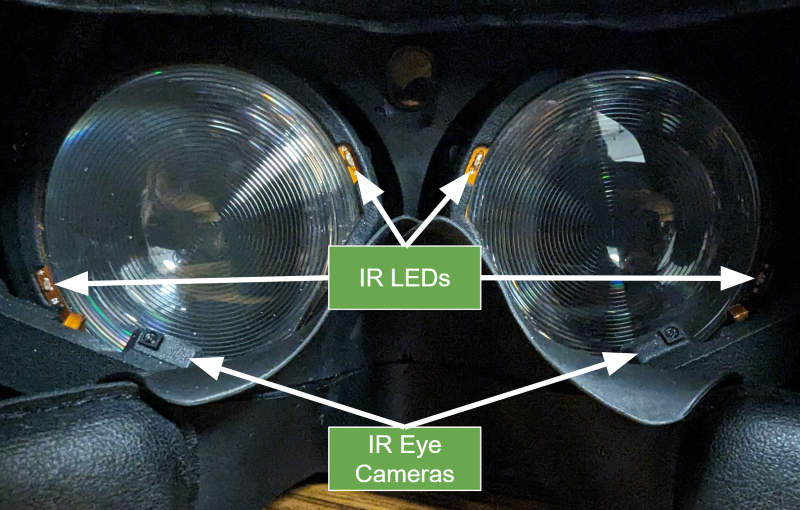}
    \caption{Pupil Labs HTC Vive Add-On, consisting of two infrared eye cameras and LEDs that fit inside the eye cavity of the HTC Vive Pro VR headset.}
    \label{fig:pl-vivepro-insert}
\end{wrapfigure}

Eye-tracking data was collected from participants using the HTC Vive Pro virtual reality headset equipped with the Pupil Labs HTC Vive Pro insert (Fig. \ref{fig:pl-vivepro-insert}). The insert captured video of each eye under illumination in the near-infrared. For each participant capture occurred at both a spatial resolution of \kb{192x192px} and a sampling rate of 200Hz, and at a spatial resolution of \kb{400x400px} and a sampling rate of 120Hz.\kb{ }The headset and insert were connected to one of two computers running the open-source Pupil Labs eye-tracking suite \cite{Pupil} for data capture. 

Pupil Capture version 3.5.14 was used to collect the data. This version was modified to facilitate automatic offline re-calibration of gaze data. The modification involves saving the 3D position of calibration targets presented to a file, so that they can be later used for the batch-processing of multiple recording sessions using multiple eye feature detection algorithms\footnote{All project-related source code will be made publicly available upon publication}. 

The data collection sequence was performed using Pupil Labs' HMD-Eyes\footnote{https://github.com/pupil-labs/hmd-eyes} integrated into Unity\footnote{https://unity.com/} version 2020.3.17f1. Gaze data, as well as data concerning the location of calibration and assessment points was exported using the Unity Experiment Framework \cite{brookes2020studying}. Each calibration and assessment point was sampled with timestamps that facilitated the post-hoc association with specific frames in the eye video of both the left and right eye.

Each participant was asked to complete four separate data collection sessions. All four sessions had their cameras' exposure settings set to "automatic" within the Pupil Labs software. Each session involved an initial process of eye tracker calibration followed by an assessment routine, both of which involved fixating on a sequential series of gaze targets presented within head-centered coordinates, and both of which are described in greater detail below. 

The calibration sequence was one of Pupil Labs' default sequences provided with their Unity plugin. The presentation sequence is automatically timed, with each target presented for 1.5 seconds against a mid-gray background.  Six calibration points were presented at 3 different depths (0.4, 0.65, and 2.0 meters), fixed in place relative to the participant's head. At each depth, 5 outer points were equally spaced around a central point, forming the shape of a pentagon. The layout of these calibration points and the multiple depths that they were rendered at were deliberately chosen to cover a broad range of the participants' field of view within the headset, as well as multiple depths in front and behind the virtual reality headset’s focal distance of 0.65 meters. 30 timestamps were recorded for each calibration point after a 0.3 second delay following its appearance in front of the subject. These timestamps would later be associated with simultaneously occurring pupil positions in each of the eye videos for the purposes of calibration. The calibration algorithms are further discussed in Section \ref{sec:gazers}.

The assessment routine involved sequential fixations at 27 assessment targets presented within a head-centered frame of reference against a mid-gray background. These targets were organized into degrees of eccentricity and were uniformly spaced in 8-point circles 20, 30, and 40 visual degrees in diameter, each with a target within the center of the visual field. All assessment targets were presented at a depth of 1 meter, and only one target was visible at a time. The assessment targets were initially presented as a black disc with a visible diameter of 1 degree that, upon trigger by the experimenter, would turn yellow for 1 second. Participants were informed that it was during this recording period that gaze data was logged to file. Participants used a hand-held audible clicking device to signal to the experimenter when they were in fixation and ready to initiate recording for that assessment target, with the instructed goal of avoiding blinks while the target was yellow. The participant would trigger recording three times for each target before the experimenter used the keyboard to transition to the next assessment target, where the process was repeated.

\subsection{Eye image segmentation and pupil detection algorithms}

Our evaluations, which were conducted using the Pupil Labs integration (developed by Pupil Labs GmbH \cite{Pupil2} in Berlin, Germany) into the HTC Vive Pro, utilize the associated Pupil Labs gaze estimation software framework. We chose to use the Pupil Labs framework because it is widely adopted, open source, and can be modified to provide tests with the appropriate controls. We chose to evaluate the influence of segmentation on the gaze estimate using the Pupil Labs HTC Vive Pro integration rather than the Pupil Labs Core tracker (mobile with a glasses-like form-factor) because the use of virtual reality facilitates the controlled presentation of gaze targets at known locations within the visual scene. Our tests utilize offline processing because many of the segmentation networks tested were not designed for real-time use, and cannot maintain the frame-rates necessary for real-time gaze estimation.

The default Pupil Labs pupil detection algorithm has several configurable parameters that can influence the quality of the pupil detection process. The first of these properties is the \textit{intensity range}, a value that determines the darkness threshold of what is considered to be a pupil. The default \textit{intensity range} is 23, which was found to be sufficient for the 192x192px eye videos. However, we found that pupil detection in the 400x400px eye videos was more accurate using an \kb{\textit{intensity range}} value of 10. The second and third properties are the \textit{pupil size min} and \textit{pupil size max}, which determine the minimum and maximum radii of what is considered to be a pupil. The defaults of these are 10px and 100px respectively, and these values were found to be sufficient for both the 192x192px and 400x400px data. Hereafter, the unadulterated Pupil Labs pupil detection algorithm will be referred to as the "\textbf{native}" algorithm.

In addition to the native algorithm, the segmentation neural networks \textit{RITnet}, \textit{EllSegGen}, and \textit{ESFnet} were used as preprocessing steps for the Pupil Labs pupil detection algorithm, as shown in (Fig. \ref{fig:pl-pipeline-flowchart}). This means that the segmentation masks produced by the neural networks were given to the Pupil Labs pupil detector instead of the raw eye image. \textit{EllSegGen} is broken down into three separate algorithms: \textit{EllSegGen}, \textit{EllSegGen (Direct Pupil)}, and \textit{EllSegGen (Direct Iris)}. These algorithms represent the use of \textit{EllSegGen} as a semantic segmentation pre-processing step for the default Pupil Labs pupil detector, the use of \textit{EllSegGen} on its own as a feature detector by extracting the pupil location directly from it, and the use of \textit{EllSegGen} on its own as a feature detector by extracting the iris location directly from it respectively. \textit{ESFnet} also has this capability, and can be broken down further into \textit{ESFnet} and \textit{ESFnet (Direct Pupil)}. This processing and the subsequent analysis were done offline rather than at the time of data collection. It should be noted that, even though we performed the data processing and analysis offline, \textit{RITnet} in particular is capable of performing in realtime (>300hz) on modern hardware due to its small model size \cite{Chaudhary2019}. \textit{RITnet} produces a segmentation mask in which each pixel of the image is assigned a group label, but was not trained to predict the parameters of the ellipse that most accurately encapsulates that pupil. \textit{EllSegGen} and \textit{ESFnet}, on the other hand, are capable of producing a per-pixel segmentation mask of the eye as well as directly predicting the parameters of the ellipse that encapsulates the pupil without the need for secondary processing with a pupil detector. For this reason, we used all three neural networks as pre-processing steps to the Pupil Labs pupil detection algorithm as well as used \textit{EllSegGen} and \textit{ESFnet} on their own as pupil detection algorithms.

The eye camera video obtained during the data collection phase was passed through a pupil detection sequence frame-by-frame in order to obtain a continuous track of the participant's pupil. In order to test the different preprocessing algorithms, the footage from each data collection session was processed multiple times, each time with a different preprocessing algorithm applied before (if applicable) passing the resulting semantic segmentation mask through the Pupil Labs pupil detection algorithm. Additionally, all data was passed independently through the native algorithm without the assistance of a segmentation network, in order to create a baseline measure to compare each algorithm against. The pupil ellipses obtained from each run of the Pupil Labs pupil detection algorithm were saved for additional analysis and gaze estimation.

The resulting ellipses of the pupil detection algorithm were fed through a custom-made processing pipeline that hooked into the open-source Pupil Labs Core software. A visual guide to this process can be seen in (Fig. \ref{fig:pl-pipeline-flowchart}). Each of the pupil ellipses produced by the pupil detection algorithm was evaluated by using a sequence of two confidence algorithms to determine the quality of the detected ellipse. The first of these was developed by Pupil Labs \cite{Kassner2014}, which scored the detected pupil on a scale from 0.0-1.0 based on how elliptical the shape was and how well the shape's edges conformed to the darkest spot in the input image. The second confidence algorithm was a threshold of the IoU between the previously detected ellipse and the current ellipse. This threshold, set at 0.98, ensured that only ellipses that closely overlapped with the previously detected ellipse would be used during the gaze calibration sequence. This was done to ensure that only fixations by the participant could be used to calibrate the gaze estimators, since the calibration sequence involved only fixations. Both confidence algorithms are capable of preventing the ellipse from being used in the calibration sequence.

\subsection[idk]{\kb{Feature}-based and \kb{3D }model-based gaze estimation algorithms}
\label{sec:gazers}
The Pupil Labs software that was used in this work comes with a \textbf{\kb{feature}-based gaze estimation algorithm} that uses a polynomial function to model the relationship between the position of the pupil centroid along the width and height of eye-image space to the x/y pixel of gaze in the world video \kb{\cite{mackworth1962head, ZHU2005124}}. This polynomial function is established using the timestamps recorded during the calibration sequence, when participants were instructed to look at a sequence of gaze targets. Since the data collection described in this work was done in virtual reality, a 640x480px 30fps virtual camera was placed at the observer's head location and used as the source for a world video. The calibration procedure, which recalls the eye video frames associated with the timestamps recorded during the presentation of each calibration point, also recalls the coordinate positions of each calibration point. By combining these and using them to establish the polynomial between the pupil locations and the calibration point locations, the Pupil Labs \kb{feature}-based gaze estimation algorithm creates a mapping between each estimated pupil position and a location on the world video (the "gaze location"). These gaze locations are then estimated for the duration of each trial.

In addition to the Pupil Labs \kb{feature}-based gaze estimation algorithm, we also test the influence of feature detection done with  \textit{RITnet}, \textit{EllSegGen}, and \textit{ESFnet} on the gaze estimate produced using Pupil Lab's open-source \textbf{\kb{3D} model-based gaze estimation algorithm} in which the production of gaze estimates relies on the initial fitting of a 3D geometric eye model within eye-camera space. \kb{\hbox{\cite{8003267}} define 3D model-based methods as those which use geometrical models of the human eye to ascertain its visual axis and estimate the gaze coordinates as points of intersection where the visual axis meets the scene. \hbox{\cite{8003267}} also identify some of the earliest examples of single-camera 3D model-based gaze estimation (\mbox{\cite{10.1007/11768029_25, 10.1145/1117309.1117349, 1634506}}). The Pupil Labs 3D model-based gaze estimation algorithm falls under this category, as it uses a single camera to position the 3D eye model given a set of detected features.} This model relies on the intuition that if the projected image of the pupil/iris boundary is tracked over time, then the systematic pupil deformations that accompany changes in eye orientation provide information that can be used to estimate the eye's center of rotation within eye camera space. Subsequently, gaze estimation is a two-step process:  first, a ray is cast from the eye camera through the image of the pupil centroid, and onto the 3D eye model.  Once the pupil has been projected onto the 3D model, a second ray is cast from the center of the 3D eyeball through the projected pupil centroid.  This 3D vector represents the gaze orientation within 3D camera space. This gaze direction in eye-space is then rotated into a direction in world space by an amount that minimizes error between gaze directions and the ground-truth 3D fixation target locations presented during the calibration sequence.

The Pupil Labs software suite relies \kb{on} an elaborated version of the Swirski model that also accounts for view-dependent refraction of the pupil by the intervening cornea and aqueous humor \cite{dierkes2019fast}. Our work specifically relies on Pupil Labs' \textit{Post-Hoc HMD 3D} implementation of this gaze mapping algorithm.  In addition to view-dependent refraction, this implementation relies on several assumptions, including a normative eye radius and a fixed geometry of the eyes relative to the eye camera inserts into the HMD. During development, we found that 3D model fits were sometimes erratic in response to high-quality pupil data. 

\kb{We took two steps to address this issue. First, we pre-fit and subsequently "froze" the 3D eye model.} During normal operation, the 3D eye model is typically updated incrementally as each pupil is segmented from the 2D eye image, each providing additional information about the 3D eye's location in eye camera space. This typically means that gaze estimates produced early in the session will be worse in quality than those produced a few minutes into the session.  Given the short duration of our capture sessions, we chose to disable incremental model updating and instead fit the 3D eye models to the entire sequence of pupil data collected during the calibration sequence before gaze direction was estimated on each frame. \kb{In addition, and in consultation with the Pupil Labs software development team, we implemented an additional filter that excluded pupils with aspect ratios exceeding Pupil Labs' recommended threshold of 0.8 from being used to update the 3D eye models. This decision was based on the feedback that pupils with aspect ratios closer to 1 provided ambiguous information with regard to eye distance, and this in-turn degraded the optimization process used to estimate the location of the eyeball's centroid. It is important to note that this filter is only applied to the samples used to fit the model used for 3D gaze estimation and not to the samples used for gaze mapping, or for the subsequent calculation of dropout rate, precision, or accuracy.}
\vspace{5mm}

\subsection{Dropout Rate, Accuracy, and Precision}
\begin{wrapfigure}{R}{0.5\textwidth}
    \vspace{-5mm}
    \includegraphics[width=\linewidth]{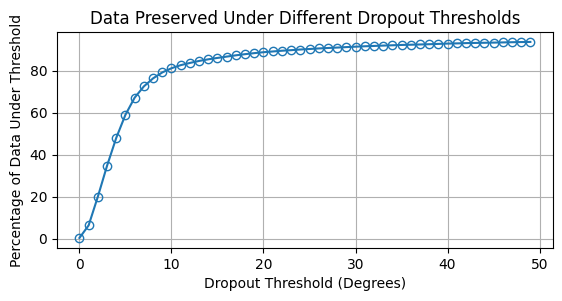}
    \caption{A comparison of dropout thresholds across all data used in the analysis portion of this experiment. Shown is the percentage of data that is retained for each dropout threshold. \kb{Due to the slope leveling out at around 10 degrees, we set the dropout threshold at 10 degrees.}}
    \label{fig:dropout-comparison}
\end{wrapfigure}
\label{sec:accandprec}

Eye feature detection will sometimes fail when a spurious eye feature has been detected where there is none, such as when a false pupil is detected in the eye lashes. \kb{Although the knowledge that this error occurred is informative on a qualitative level, the quantitative magnitude of the error is not informative of the quality of the pupil detection algorithm. For this reason, we have opted to drop these samples from subsequent calculations of accuracy and precision using a fixed dropout threshold, as in \mbox{ \cite{Macinnes299925}}. A comparison of dropout thresholds is presented in \mbox{Fig. \ref{fig:dropout-comparison}}, which presents the cumulative amount of gaze data that lies within the bounds of thresholds ranging from 0-50 degrees of accuracy error in the gaze estimate. The dropout threshold was set to 10 degrees of accuracy error on the basis that approximately 80\mbox{\%} of the gaze data lies  under the threshold, and because the change in cumulative gaze data above that threshold is very gradual over the range of 10-40 degrees of error. Gaze estimates equal to or above this threshold contribute to the dropout rate metric and were removed from further analysis.}

Accuracy represents the distance between the centroid of all gaze angles collected during fixation at a single assessment target and the ground-truth target location, both defined units of visual degrees along the azimuth and elevation, consistent with the formula: 
\begin{equation*}
    err_{acc} = \frac{\sum^n_i\sqrt{(\bar{a} - a_i)^2 + (\bar{e} - e_i)^2}}{n}
\end{equation*}

\noindent where $n$ is the number of estimated gaze locations in the group, $a_i$ and $e_i$ are the azimuth and elevation of the group's $i$th estimated gaze location in the head-centered spherical reference frame respectively, and $\bar{a}$ and $\bar{e}$ are the azimuth and elevation of the associated known ground truth fixation point in the head-centered spherical reference frame.

The precision error for a fixation point is defined as the average difference between each calculated gaze location in the fixation and the center of mass of all gaze locations sampled during the fixation. The formula for the precision error of a single fixation is shown below,

\begin{equation*}
    \begin{aligned}
        \mu^a &= \frac{\sum^n_ia_i}{n}\;\;\;\;\;\;\;
        \mu^e = \frac{\sum^n_ie_i}{n}\\
        err_{prec} &= \frac{\sum^n_i\sqrt{(\mu^a - a_i)^2 + (\mu^e - e_i)^2}}{n}\\
    \end{aligned}
\end{equation*}

\noindent where \textit{n} is the number of estimated gaze locations in the group, and $a_i$ and $e_i$ are the azimuth and elevation of the group's $i$th estimated gaze location in the head-centered spherical reference frame respectively.

\section{Results}
\begin{figure}
    \includegraphics[width=\linewidth]{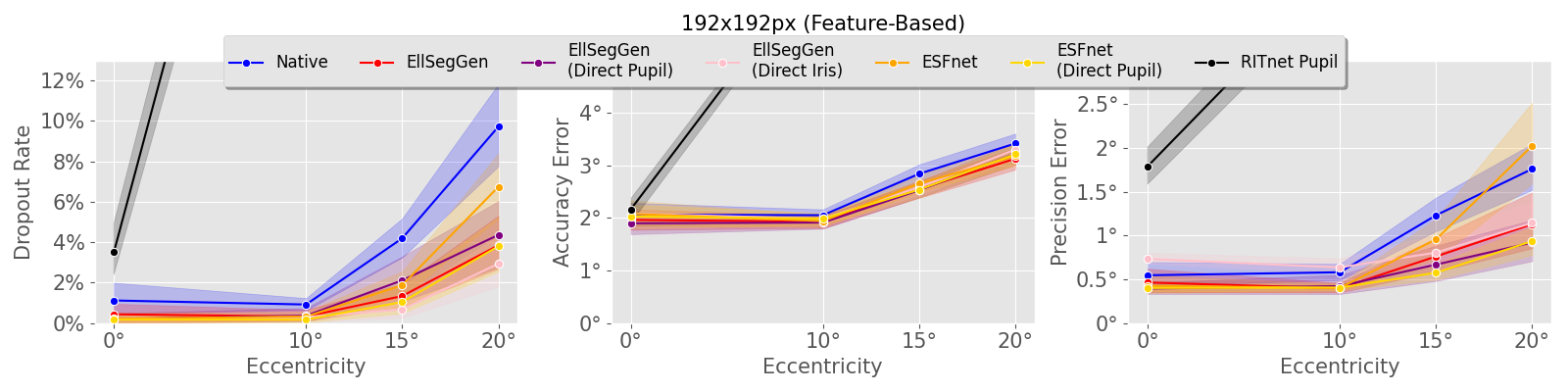}
    \includegraphics[width=\linewidth]{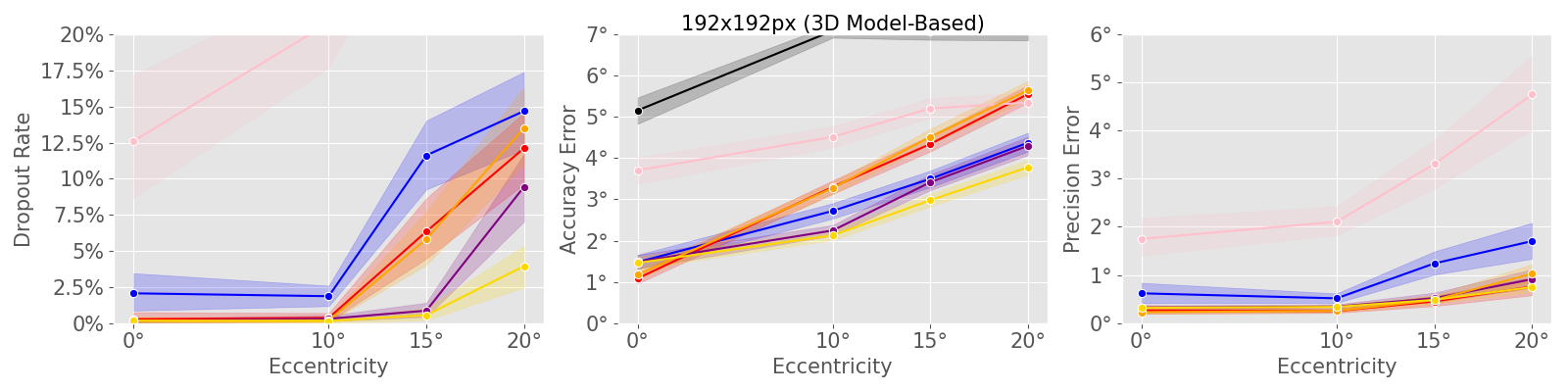}
    \caption{Dropout rate (left), accuracy error (center), and precision error (right) across fixation point eccentricities for the 192x192px eye data collected from the \kb{feature}-based (top) and \kb{3D} model-based (bottom) gaze estimation algorithms. Samples above the dropout threshold of 10$^{\circ}$ were omitted from calculations of accuracy and precision. Shading represents 95\% confidence intervals for the mean. The range of the Y axis were chosen to provide insight into the performance of the best-performing algorithms, with the consequence that \textit{RITnet}'s and \textit{EllSegGen} (Direct Iris)'s error falls beyond its range in some graphs. This data is presented in Tables \ref{tab:robustness}, \ref{tab:accuracy}, \ref{tab:precision}.}
    \label{fig:ecc-separated-192}
\end{figure}

\begin{figure}
    \includegraphics[width=\linewidth]{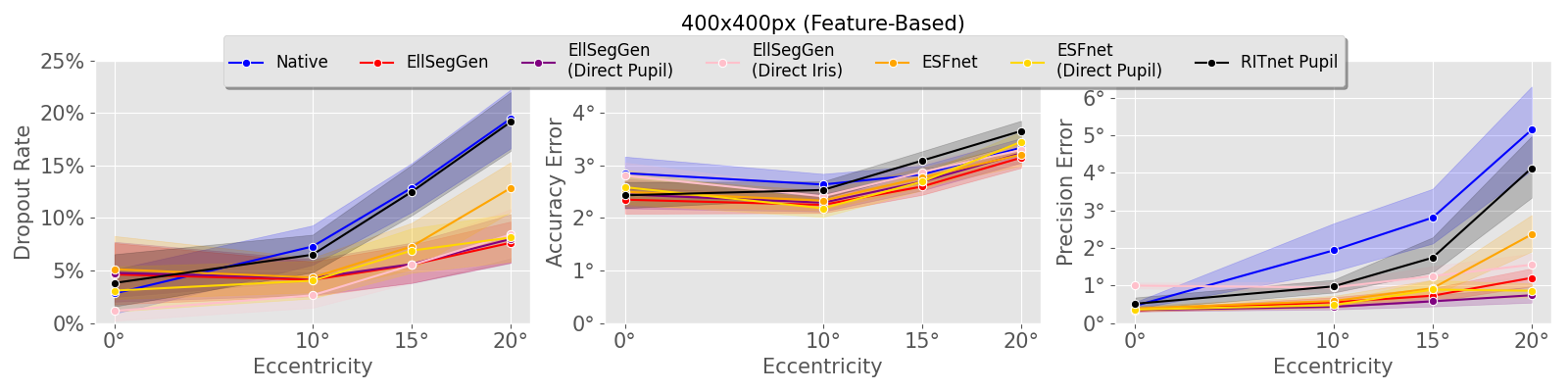}
    \includegraphics[width=\linewidth]{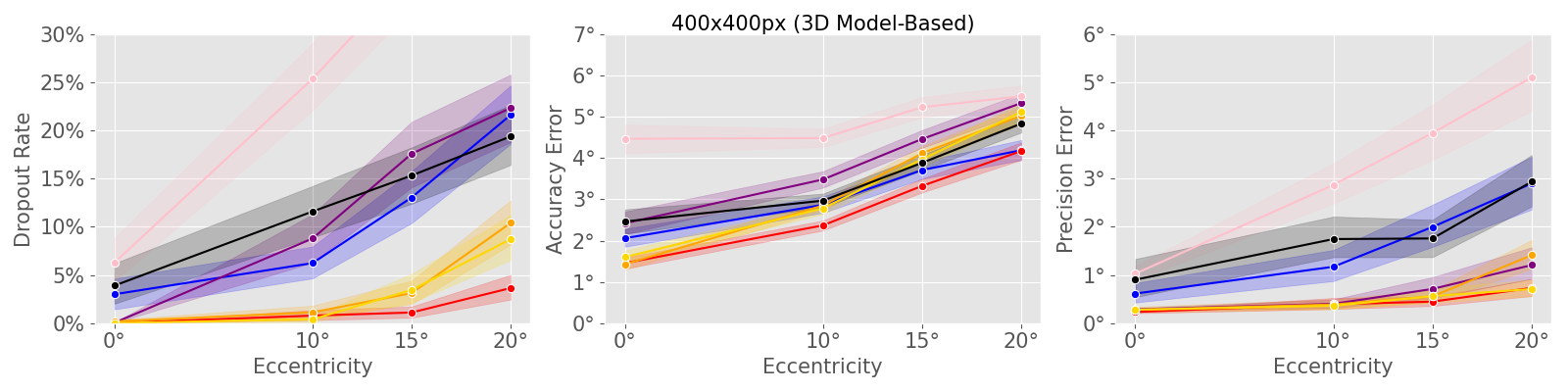}
    \caption{Dropout rate (left), accuracy error (center), and precision error (right) across fixation point eccentricities for the 400x400px eye data collected from the \kb{feature}-based (top) and \kb{3D} model-based (bottom) gaze estimation algorithms. Samples above the dropout threshold of 10$^{\circ}$ were omitted from calculations of accuracy and precision. Shading represents 95\% confidence intervals for the mean. This data is presented in numerical form in Tables \ref{tab:robustness}, \ref{tab:accuracy}, \ref{tab:precision}.}
    \label{fig:ecc-separated-400}
\end{figure}

Results, averaged across all participants for dropout rate, accuracy, and precision are in the left, center, and right columns respectively of Fig. \ref{fig:ecc-separated-192} (for 192 x 192px images) and Fig. \ref{fig:ecc-separated-400} (for 400 x 400px images). Additionally, the mean and standard error across all participants for dropout rate, accuracy, and precision are shown in Tables \ref{tab:robustness}, \ref{tab:accuracy}, and  \ref{tab:precision} respectively.

\subsection{RITnet}
\begin{wraptable}{r}{0.45\textwidth}\centering
\vspace{-4mm}
    \scriptsize
    \resizebox{\linewidth}{!}{%
    \begin{tabular}{l|rr|rrr}\toprule
        \ul{\textbf{Dropout Rate (192)}} &\ul{\textbf{Feature}} & &\ul{\textbf{3D Model}} & \\
        \ul{\textbf{}} &\ul{Mean} &\ul{Std. Error} &\ul{Mean} &\ul{Std. Error} \\
        Native &4.52\% &1.52\% &8.58\% &5.03\% \\
        EllSegGen &1.68\% &0.77\% &5.61\% &3.82\% \\
        EllSegGen (Direct Pupil) &2.05\% &1.47\% &3.16\% &1.66\% \\
        EllSegGen (Direct Iris) &\cellcolor[HTML]{b7e1cd}1.18\% &\cellcolor[HTML]{b7e1cd}0.66\% &\cellcolor[HTML]{ff6347}30.45\%* &\cellcolor[HTML]{ff6347}8.15\%* \\
        ESFnet &2.67\% &0.96\% &5.80\% &3.29\% \\
        ESFnet (Direct Pupil) &1.51\% &0.72\% &\cellcolor[HTML]{b7e1cd}1.39\% &\cellcolor[HTML]{b7e1cd}1.03\% \\
        RITnet (Pupil) &61.57\% &2.65\% &81.16\% &2.44\% \\
        \ul{\textbf{Dropout Rate (400)}} &\ul{\textbf{Feature}} & &\ul{\textbf{3D Model}} & \\
        &\ul{Mean} &\ul{Std. Error} &\ul{Mean} &\ul{Std. Error} \\
        Native &12.05\% &4.67\% &12.45\% &4.95\% \\
        EllSegGen &5.65\% &3.97\% &\cellcolor[HTML]{b7e1cd}1.64\% &\cellcolor[HTML]{b7e1cd}1.05\% \\
        EllSegGen (Direct Pupil) &5.84\% &4.16\% &14.44\% &6.64\% \\
        EllSegGen (Direct Iris) &\cellcolor[HTML]{b7e1cd}5.05\% &\cellcolor[HTML]{b7e1cd}3.45\% &\cellcolor[HTML]{ff6347}32.46\%* &9.20\%\cellcolor[HTML]{ff6347}* \\
        ESFnet &7.82\% &4.73\% &4.36\% &2.45\% \\
        ESFnet (Direct Pupil) &6.00\% &4.17\% &3.73\% &1.58\% \\
        RITnet (Pupil) &11.72\% &5.66\% &14.16\% &5.81\% \\
        \bottomrule
    \end{tabular}}
    \caption{Dropout rate mean values across subjects for the \kb{feature}-based and \kb{3D} model-based gaze estimators. Shaded green are the best-scoring values for the category (lower is better). This data encompasses the 192x192px resolution (top) and the 400x400px resolution (bottom). *\textit{EllSegGen (Direct Iris)}, when used in conjunction with the \kb{3D} model-based eye tracker, is not recommended. See Section \ref{sec:discussion}.}
    \label{tab:robustness}
    %\vspace{-15mm}
\end{wraptable}

At the \kb{192x192px} resolution, \textit{RITnet} performed the poorest overall in terms of dropout, accuracy error, and precision error compared to the other detection methods. Average dropout rates were above 61\% for \kb{feature}-based methods and above 81\% for \kb{3D} model-based methods (see Table \ref{tab:robustness}). Accuracy and precision were similarly poor, and together these results indicate that \textit{RITnet} is unsuitable for use at the 192x192 resolution. 

At the 400x400px eye image resolution, \textit{RITnet} demonstrated performance that was largely equivalent to that of the native Pupil Labs algorithm. Like the native Pupil Labs algorithm, dropout rate increased approximately linearly with the eccentricity  of gaze angles, from $\sim$5\% to a peak of $\sim$20\% when passed through \kb{feature}-based and \kb{3D} model-based gaze estimation algorithms. Accuracy error remained within 0.5 degrees of the native algorithm for both the \kb{feature}-based and \kb{3D} model-based gaze estimators. There was some improvement to the dropout rate and precision over the native algorithm when passed through the \kb{feature}-based gaze estimation algorithm, though this improvement can be attributed to only three of ten individuals. When using the \kb{3D} model-based gaze estimation algorithm, the output from the native pupil detector improved to match that of \textit{RITnet}.\\

 \subsection{EllSegGen}
 \label{sec:results-ellseg}

When output from \textit{EllSegGen (Direct Iris)} was passed through the \kb{3D} model-based gaze estimation algorithm, dropout rates were so high (30\% for 192x192, 32\% for 400x400, see Table \ref{tab:robustness}) as to not be recommended compared to the other \textit{EllSegGen} options. For this reason, the use of \textit{EllsegGen (Direct Iris)} in combination with \kb{3D} model-based methods will not receive further discussion in this section. However, the issue will be revisited in the discussion (Section \ref{sec:discussion}).

At an eye image resolution of 192x192, \textit{EllSegGen} and \textit{EllSegGen (Direct Pupil)} demonstrated noticeably lower dropout rates than the native algorithm across all eccentricities, for both gaze estimation types. \textit{EllSegGen (Direct Iris)} demonstrated similarly lowered dropout rates than the native algorithm for the \kb{feature}-based gaze estimation algorithm, outperforming all other algorithms in the category. However, there was no improvement to accuracy over the native algorithm when using \kb{feature}-based gaze estimation algorithms, and performance was equivalent to or worse than that of the native algorithm when features were then passed through a \kb{3D} model-based estimation algorithm. There were modest improvements to precision when compared to the native algorithm for both \kb{feature} and \kb{3D} model-based estimation algorithms.
\begin{wraptable}{l}{0.45\textwidth}\centering
    \vspace{2mm}
    \scriptsize
    \resizebox{\linewidth}{!}{%
    \begin{tabular}{l|rr|rrr}\toprule
        \ul{\textbf{Accuracy Error (192)}} &\ul{\textbf{Feature}} & &\ul{\textbf{3D Model}} & \\
        \ul{\textbf{}} &\ul{Mean} &\ul{Std. Error} &\ul{Mean} &\ul{Std. Error} \\
        Native &2.691 &0.310 &3.375 &0.494 \\
        EllSegGen &\cellcolor[HTML]{b7e1cd}2.471 &0.345 &4.048 &0.382 \\
        EllSegGen (Direct Pupil) &2.483 &0.306 &3.124 &0.421 \\
        EllSegGen (Direct Iris) &2.563 &0.277 &\cellcolor[HTML]{ff6347}4.962* &\cellcolor[HTML]{ff6347}0.542* \\
        ESFnet &2.534 &0.340 &4.115 &0.377 \\
        ESFnet (Direct Pupil) &2.519 &0.293 &\cellcolor[HTML]{b7e1cd}2.802 &0.345 \\
        RITnet (Pupil) &5.969 &\cellcolor[HTML]{b7e1cd}0.143 &6.771 &\cellcolor[HTML]{b7e1cd}0.080 \\
        \ul{\textbf{Accuracy Error (400)}} &\ul{\textbf{Feature}} & &\ul{\textbf{3D Model}} & \\
        &\ul{Mean} &\ul{Std. Error} &\ul{Mean} &\ul{Std. Error} \\
        Native &3.053 &0.468 &3.444 &0.525 \\
        EllSegGen &\cellcolor[HTML]{b7e1cd}2.700 &0.340 &\cellcolor[HTML]{b7e1cd}3.084 &0.369 \\
        EllSegGen (Direct Pupil) &2.775 &\cellcolor[HTML]{b7e1cd}0.309 &4.335 &0.442 \\
        EllSegGen (Direct Iris) &2.869 &0.335 &\cellcolor[HTML]{ff6347}5.145* &\cellcolor[HTML]{ff6347}0.567* \\
        ESFnet &2.836 &0.399 &3.713 &0.352 \\
        ESFnet (Direct Pupil) &2.833 &0.360 &3.684 &\cellcolor[HTML]{b7e1cd}0.299 \\
        RITnet (Pupil) &3.084 &0.384 &3.814 &0.431 \\
        \bottomrule
    \end{tabular}}
    \caption{Accuracy error mean values (degrees) across subjects for the \kb{feature}-based and \kb{3D} model-based gaze estimators after dropouts have been filtered from the data. Shaded green are the best-scoring values for the category (lower is better). This data encompasses the 192x192px resolution (top) and 400x400px resolution (bottom). *\textit{EllSegGen (Direct Iris)}, when used in conjunction with the \kb{3D} model-based eye tracker, is not recommended. See Section \ref{sec:discussion}.}
    \label{tab:accuracy}
    \vspace{-8mm}
\end{wraptable}
\begin{wraptable}{r}{0.45\textwidth}\centering
    \vspace{-6mm}
    \scriptsize
    \resizebox{\linewidth}{!}{%
    \begin{tabular}{l|rr|rrr}\toprule
        \ul{\textbf{Precision Error (192)}} &\ul{\textbf{Feature}} & &\ul{\textbf{3D Model}} & \\
        \ul{\textbf{}} &\ul{Mean} &\ul{Std. Error} &\ul{Mean} &\ul{Std. Error} \\
        Native &1.117 &0.277 &1.184 &0.430 \\
        EllSegGen &0.729 &0.231 &\cellcolor[HTML]{b7e1cd}0.502 &0.172 \\
        EllSegGen (Direct Pupil) &0.643 &0.227 &0.563 &0.118 \\
        EllSegGen (Direct Iris) &0.847 &0.168 &\cellcolor[HTML]{ff6347}3.291* &\cellcolor[HTML]{ff6347}0.774* \\
        ESFnet &1.047 &0.335 &0.565 &0.153 \\
        ESFnet (Direct Pupil) &\cellcolor[HTML]{b7e1cd}0.613 &\cellcolor[HTML]{b7e1cd}0.126 &0.503 &\cellcolor[HTML]{b7e1cd}0.087 \\
        RITnet (Pupil) &6.346 &0.442 &12.921 &1.185 \\
        \ul{\textbf{Precision Error (400)}} &\ul{\textbf{Feature}} & &\ul{\textbf{3D Model}} & \\
        &\ul{Mean} &\ul{Std. Error} &\ul{Mean} &\ul{Std. Error} \\
        Native &2.885 &1.100 &1.913 &0.690 \\
        EllSegGen &0.800 &0.245 &\cellcolor[HTML]{b7e1cd}0.491 &0.179 \\
        EllSegGen (Direct Pupil) &\cellcolor[HTML]{b7e1cd}0.589 &\cellcolor[HTML]{b7e1cd}0.176 &0.735 &0.204 \\
        EllSegGen (Direct Iris) &1.242 &0.382 &\cellcolor[HTML]{ff6347}3.927* &\cellcolor[HTML]{ff6347}1.070* \\
        ESFnet &1.257 &0.417 &0.729 &0.280 \\
        ESFnet (Direct Pupil) &0.730 &0.238 &0.508 &\cellcolor[HTML]{b7e1cd}0.155 \\
        RITnet (Pupil) &2.228 &0.836 &2.140 &0.721 \\
        \bottomrule
    \end{tabular}}
    \caption{Precision error mean values (degrees) across subjects for the \kb{feature}-based and \kb{3D} model-based gaze estimators after dropouts have been filtered from the data. Shaded green are the best-scoring values for the category (lower is better). This data encompasses the 192x192px resolution (top) and 400x400px resolution (bottom). *\textit{EllSegGen (Direct Iris)}, when used in conjunction with the \kb{3D} model-based eye tracker, is not recommended. See Section \ref{sec:discussion}.}
    \label{tab:precision}
    \vspace{-14mm}
\end{wraptable}

At the \kb{400x400px} resolution, in combination with \kb{feature}-based methods, \textit{EllSegGen} and its two variants have mean dropout rates below 6\% with little variation between eccentricities. Although the accuracy error does not show consistent improvement over the native algorithm across different eccentricities, there are noticeable improvements to precision. The improvement is negligible at 0$^{\circ}$ eccentricity, but increases to $\sim$4$^{\circ}$ at 20$^{\circ}$ eccentricity. \textit{EllSegGen (Direct Iris)} demonstrates similar improvements for \kb{feature}-based methods, with only a modest increase in precision error at the highest eccentricity. In contrast, \textit{EllSegGen} is the best-performing option when gaze is estimated using \kb{3D} model-based methods at a 400x400px resolution. Dropout rates remain below 5\% at all eccentricities, and accuracy remains lowest at all eccentricities, although the improvement over the native algorithm is still modest. The most notable improvement, however, is that precision remains below 1.5$^{\circ}$ across all eccentricities.

\vspace{-3.5mm}
\subsection{ESFnet}

At an eye image resolution of 192x192px, \textit{ESFnet} performed similarly to \textit{EllSegGen}. The one exception is a drop in precision at greater eccentricities using \kb{feature}-based methods. In contrast, \textit{ESFnet (Direct Pupil)} matches or outperforms all other models in every metric, with the exception of dropout rate, which increased with eccentricity when using \kb{feature}-based gaze estimation methods. % Nevertheless, rates were lower than 7\% at their highest, and these rates are likely negligible in most use cases.

When operating on the 400x400px data, \textit{ESFnet} and \textit{ESFnet (Direct Pupil)} match or exceed the performance of the native algorithm when passed through either \kb{feature} or \kb{3D} model-based gaze estimation methods. Their performance is very similar to \textit{EllSegGen} in every metric, with the one notable exception that there is an increase in dropout rate at greater eccentricities when passed through \kb{3D} model-based methods.

\section{Discussion}
\label{sec:discussion}

\begin{wrapfigure}{R}{0.55\textwidth}
    \vspace{-0.25cm}
    \includegraphics[width=0.55\textwidth]{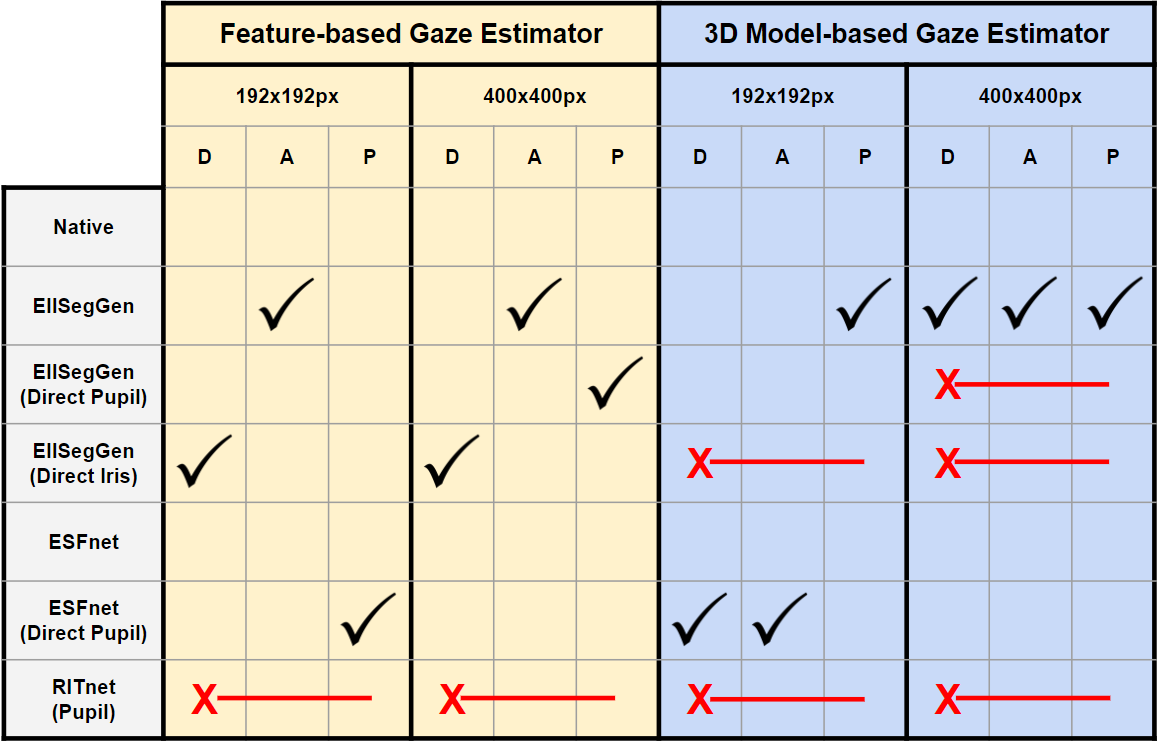}
    \caption{Summary of best-performing feature detection techniques in each category (indicated with a checkmark). \textbf{D: Dropout Rate; A: Accuracy; P: Precision}.       Since dropout rates have a cascading effect on both accuracy and precision, models with a higher dropout rate than the native approach are marked with a red \textit{X}. Regardless of the numerical performance of the accuracy and precision, we do not recommend using models with a high dropout rate -- indicated by a red line through the remaining categories. }
    \label{fig:bench}
\end{wrapfigure}

Fig. \ref{fig:bench} summarizes the best options in terms of dropout rate, accuracy, and precision. Nonetheless, care must be taken when interpreting these results. In particular, a high dropout rate means fewer samples are used to compute the accuracy and precision metrics which could potentially lead to skewed results. 

The positive performance demonstrated by the ML segmentation models tested here is encouraging when one considers that the application context was not well-represented in any of the datasets used to train these models. In particular, the eye images passed into the models in our tests were taken from off-axial angles within a virtual reality headset using the Pupil Labs HTC Vive Add-On. In contrast, the \textit{OpenEDS} dataset \cite{Garbin2019}, which all tested segmentation models were trained on, contains eye images taken on-axis inside a virtual reality headset. The Labeled Pupils in Wild (LPW) \cite{Tonsen2015} dataset, which \textit{EllSegGen} and \textit{ESFnet} were trained on, contains eye images captured from a mobile eye tracker in real-world environments outside of virtual reality, both indoors and outdoors. The \textit{Swirski} dataset \cite{Swirski2012}, which was used in training \textit{EllSegGen}, also contained eye images captured from an eye tracker outside of virtual reality. The BAT dataset was also used to train \textit{EllSegGen}, and is composed of eye images from physically-restrained subjects rather than subjects in virtual reality. The last of the non-artificial datasets used by both \textit{EllSegGen} and \textit{ESFnet} are the \textit{Fuhl} datasets, defined as a blend of the \textit{ExCuSe} \cite{Fuhl2015a}, \textit{ElSe} \cite{Fuhl2015b}, and \textit{PupilNet} \cite{Fuhl2016b} datasets. These datasets contain eye images captured from mobile eye trackers while performing tasks such as driving. In addition, multiple datasets containing artificial eye images were used in the training of \textit{EllSegGen} and \textit{ESFnet} \cite{Nair2020, Kim2019, Nair2020thesis, wood2016_etra}. This demonstrates that these machine learning models are capable of generalizing to mobile eye trackers with parameters that differ from the datasets used to train them.

\kb{Despite the positive impact of ML models, image resolution had a strong effect on the performance of several models, with the counter-intuitive result that models applied to the lower resolution 192x192px images outperformed their application to the higher resolution 400x400px images. We attributed this result to one of two possibilities: to a difference between the resolution of training images and the images collected by the eye tracker in our tests, or to confounding differences in image content across the 192x192px and 400x400px datasets used in our study. To test these competing hypotheses, we took a high-resolution dataset (OpenEDS) in which images are provided along-side ground-truth pupil segmentation masks. We then took each of our detector plugins and measured the error between the centroids of our algorithm-segmented pupils and the centroids of the ground truth pupils. This measure was taken using both cropped 400x400px images and corresponding versions that had been down-sampled to 192x192px, but that were otherwise identical. To account for differences in resolution, error was measured in units of image width. This test revealed that performance was no better on the 192x192px than 400x400px pixel variants. The observation that the effect of resolution disappeared when image content was held constant suggests that the effect of image resolution in our study can be attributed to differences in image content between the 192x192px and 400x400px images.}

The findings also reveal that \textit{EllSegGen (Direct Iris)} can provide very competitive results when passed through a \kb{feature}-based gaze estimator, but disastrous results when passed through a \kb{3D} model-based gaze estimator. This likely reflects inherent assumptions in the Pupil Labs \textit{Pye3D} \kb{3D} model-based gaze estimator about pupil size that have not been well documented, but that prevent iris features from being used to estimate the centroid of the eyeball. Alternatively, this may reflect differences in the information that the two estimators rely on to produce their estimates. Whereas the \kb{feature}-based gaze estimation algorithm only utilizes the centroid of the detected ellipse, the \kb{3D} model-based algorithm uses all of the ellipse's parameters to position the 3D eye models. As a result, any flaws in the model's ability to accurately estimate all of the parameters of the iris ellipse would have a more significant effect on the \kb{3D} model-based algorithm than on the \kb{feature}-based algorithm. This raises the possibility that \textit{EllSegGen (Direct Iris)} is very good at detecting the iris centroid while also being very bad at estimating the iris boundaries. 

\section{Conclusion}

The large disparity in data quality between remote and mobile eye trackers has made attempts to explore gaze behavior in experimental contexts outside the laboratory difficult. The hardware constraints necessitated by the form factor and head-mounted nature of a mobile eye tracker often result in output data of limited quality. Therefore, in order to maximize the quality of mobile eye tracker data, it becomes necessary to apply sophisticated software techniques to the data prior to performing more traditional eye tracking algorithms.

The most commonly used consumer-level mobile eye tracking systems have not taken advantage of machine learning for accurate eye image segmentation or feature localization. In this paper, we have measured the potential effect that machine learning improvements to the accuracy of the feature localization stage could have on the quality of the final gaze estimate. These measurements have led us to conclude that high-performing eye feature detection neural networks are capable of improving the dropout rate and precision of gaze estimates without negatively affecting the gaze accuracy. \kb{Hence, our work provides users of eye tracking systems a means to reduce dropouts in their own data through the informed selection of pupil detection models.}

This manuscript has discussed the groundwork we have laid for evaluating the contribution of feature detection models on the quality of the gaze estimate when applied to a widely adopted open-source eye tracking solution. The software we have written and used to evaluate these contributions, including the software pipeline for the controlled evaluation of the feature-detection stage and the additional modules for the evaluation of several contemporary eye segmentation networks, will be made publicly available upon publication of this manuscript.

Though our evaluations rely on the Pupil Labs integration into the HTC Vive Pro headset and the associated Pupil Labs gaze estimation software framework, none of the findings of this paper are necessarily dependent on either of these factors. Future work may evaluate the viability of these machine learning solutions on alternative eye-tracking hardware/software, and outside of virtual reality. Outdoor eye tracking, especially in direct sunlight, is particularly challenging \cite{Binaee2021} and may benefit significantly from the use of state-of-the-art feature detection neural networks. It is also possible that further contributions to this area, especially on the side of the machine learning algorithms, may result in a machine learning-assisted eye tracking pipeline that is capable of running in real-time.

\begin{acks}
We would like to thank Dr. Aayush Chaudhary and Dr. Rakshit Kothari for their feedback and guidance on the use of RITnet and EllSeg. We would also like Dr. Jeff Pelz for his mentorship and advice over the course of this project. Finally, we extend our thanks to Pablo Prietz for his helpful insights on the inner workings of the Pupil Labs Core pipeline.
\end{acks}

\bibliographystyle{ACM-Reference-Format}
\bibliography{main}


\begin{thebibliography}{39}

%%% the Web address.

\ifx \showCODEN    \undefined \def \showCODEN     #1{\unskip}     \fi
\ifx \showDOI      \undefined \def \showDOI       #1{#1}\fi
\ifx \showISBNx    \undefined \def \showISBNx     #1{\unskip}     \fi
\ifx \showISBNxiii \undefined \def \showISBNxiii  #1{\unskip}     \fi
\ifx \showISSN     \undefined \def \showISSN      #1{\unskip}     \fi
\ifx \showLCCN     \undefined \def \showLCCN      #1{\unskip}     \fi
\ifx \shownote     \undefined \def \shownote      #1{#1}          \fi
\ifx \showarticletitle \undefined \def \showarticletitle #1{#1}   \fi
\ifx \showURL      \undefined \def \showURL       {\relax}        \fi
% invisible to TeX
\providecommand\bibfield[2]{#2}
\providecommand\bibinfo[2]{#2}
\providecommand\natexlab[1]{#1}
\providecommand\showeprint[2][]{arXiv:#2}

\bibitem[Binaee et~al\mbox{.}(2021)]%
        {Binaee2021}
\bibfield{author}{\bibinfo{person}{Kamran Binaee}, \bibinfo{person}{Christian Sinnott}, \bibinfo{person}{Kaylie~Jacleen Capurro}, \bibinfo{person}{Paul MacNeilage}, {and} \bibinfo{person}{Mark~D Lescroart}.} \bibinfo{year}{2021}\natexlab{}.
\newblock \showarticletitle{Pupil Tracking Under Direct Sunlight}. In \bibinfo{booktitle}{\emph{ACM Symposium on Eye Tracking Research and Applications}} (Virtual Event, Germany) \emph{(\bibinfo{series}{ETRA '21 Adjunct})}. \bibinfo{publisher}{Association for Computing Machinery}, \bibinfo{address}{New York, NY, USA}, Article \bibinfo{articleno}{18}, \bibinfo{numpages}{4}~pages.
\newblock
\showISBNx{9781450383578}
\urldef\tempurl%
\url{https://doi.org/10.1145/3450341.3458490}
\showDOI{\tempurl}


\bibitem[Brookes et~al\mbox{.}(2020)]%
        {brookes2020studying}
\bibfield{author}{\bibinfo{person}{Jack Brookes}, \bibinfo{person}{Matthew Warburton}, \bibinfo{person}{Mshari Alghadier}, \bibinfo{person}{Mark Mon-Williams}, {and} \bibinfo{person}{Faisal Mushtaq}.} \bibinfo{year}{2020}\natexlab{}.
\newblock \showarticletitle{Studying human behavior with virtual reality: The Unity Experiment Framework}.
\newblock \bibinfo{journal}{\emph{Behavior research methods}}  \bibinfo{volume}{52} (\bibinfo{year}{2020}), \bibinfo{pages}{455--463}.
\newblock


\bibitem[Cai et~al\mbox{.}(2021)]%
        {Cai2021}
\bibfield{author}{\bibinfo{person}{Xin Cai}, \bibinfo{person}{Jiabei Zeng}, {and} \bibinfo{person}{Shiguang Shan}.} \bibinfo{year}{2021}\natexlab{}.
\newblock \showarticletitle{Landmark-aware Self-supervised Eye Semantic Segmentation}.
\newblock \bibinfo{journal}{\emph{Proceedings - 2021 16th IEEE International Conference on Automatic Face and Gesture Recognition, FG 2021}} (\bibinfo{year}{2021}).
\newblock
\showISBNx{9781665431767}
\urldef\tempurl%
\url{https://doi.org/10.1109/FG52635.2021.9667031}
\showDOI{\tempurl}


\bibitem[Chaudhary(2019)]%
        {Chaudhary2019b}
\bibfield{author}{\bibinfo{person}{Aayush~K. Chaudhary}.} \bibinfo{year}{2019}\natexlab{}.
\newblock \showarticletitle{Motion Tracking of Iris Features for Eye Tracking}. In \bibinfo{booktitle}{\emph{Proceedings of the 11th ACM Symposium on Eye Tracking Research \&amp; Applications}} (Denver, Colorado) \emph{(\bibinfo{series}{ETRA '19})}. \bibinfo{publisher}{Association for Computing Machinery}, \bibinfo{address}{New York, NY, USA}, Article \bibinfo{articleno}{53}, \bibinfo{numpages}{3}~pages.
\newblock
\showISBNx{9781450367097}
\urldef\tempurl%
\url{https://doi.org/10.1145/3314111.3322872}
\showDOI{\tempurl}


\bibitem[Chaudhary et~al\mbox{.}(2019)]%
        {Chaudhary2019}
\bibfield{author}{\bibinfo{person}{Aayush~K. Chaudhary}, \bibinfo{person}{Rakshit Kothari}, \bibinfo{person}{Manoj Acharya}, \bibinfo{person}{Shusil Dangi}, \bibinfo{person}{Nitinraj Nair}, \bibinfo{person}{Reynold Bailey}, \bibinfo{person}{Christopher Kanan}, \bibinfo{person}{Gabriel Diaz}, {and} \bibinfo{person}{Jeff~B. Pelz}.} \bibinfo{year}{2019}\natexlab{}.
\newblock \showarticletitle{RITnet: Real-time Semantic Segmentation of the Eye for Gaze Tracking}.
\newblock \bibinfo{journal}{\emph{Proceedings - 2019 International Conference on Computer Vision Workshop, ICCVW 2019}} (\bibinfo{date}{10} \bibinfo{year}{2019}), \bibinfo{pages}{3698--3702}.
\newblock
\urldef\tempurl%
\url{https://doi.org/10.1109/iccvw.2019.00568}
\showDOI{\tempurl}
\newblock
\shownote{RITnet Original Paper}.


\bibitem[Dierkes et~al\mbox{.}(2019)]%
        {dierkes2019fast}
\bibfield{author}{\bibinfo{person}{Kai Dierkes}, \bibinfo{person}{Moritz Kassner}, {and} \bibinfo{person}{Andreas Bulling}.} \bibinfo{year}{2019}\natexlab{}.
\newblock \showarticletitle{A fast approach to refraction-aware eye-model fitting and gaze prediction}.
\newblock \bibinfo{journal}{\emph{Eye Tracking Research and Applications Symposium (ETRA)}} (\bibinfo{date}{6} \bibinfo{year}{2019}).
\newblock
\showISBNx{9781450367097}
\urldef\tempurl%
\url{https://doi.org/10.1145/3314111.3319819}
\showDOI{\tempurl}


\bibitem[Fuhl et~al\mbox{.}(2015a)]%
        {Fuhl2015a}
\bibfield{author}{\bibinfo{person}{Wolfgang Fuhl}, \bibinfo{person}{Thomas Kübler}, \bibinfo{person}{Katrin Sippel}, \bibinfo{person}{Wolfgang Rosenstiel}, {and} \bibinfo{person}{Enkelejda Kasneci}.} \bibinfo{year}{2015}\natexlab{a}.
\newblock \showarticletitle{ExCuSe: Robust Pupil Detection in Real-World Scenarios}.
\newblock \bibinfo{journal}{\emph{Lecture Notes in Computer Science (including subseries Lecture Notes in Artificial Intelligence and Lecture Notes in Bioinformatics)}}  \bibinfo{volume}{9256} (\bibinfo{year}{2015}), \bibinfo{pages}{39--51}.
\newblock
\showISBNx{9783319231914}
\showISSN{16113349}
\urldef\tempurl%
\url{https://doi.org/10.1007/978-3-319-23192-1_4}
\showDOI{\tempurl}


\bibitem[Fuhl et~al\mbox{.}(2016a)]%
        {Fuhl2016b}
\bibfield{author}{\bibinfo{person}{Wolfgang Fuhl}, \bibinfo{person}{Thiago Santini}, \bibinfo{person}{Gjergji Kasneci}, {and} \bibinfo{person}{Enkelejda Kasneci}.} \bibinfo{year}{2016}\natexlab{a}.
\newblock \showarticletitle{PupilNet: Convolutional Neural Networks for Robust Pupil Detection}.
\newblock  (\bibinfo{date}{1} \bibinfo{year}{2016}).
\newblock
\showISBNx{1601.04902v1}
\urldef\tempurl%
\url{https://arxiv.org/abs/1601.04902v1}
\showURL{%
\tempurl}


\bibitem[Fuhl et~al\mbox{.}(2015b)]%
        {Fuhl2015b}
\bibfield{author}{\bibinfo{person}{Wolfgang Fuhl}, \bibinfo{person}{Thiago~C. Santini}, \bibinfo{person}{Thomas Kübler}, {and} \bibinfo{person}{Enkelejda Kasneci}.} \bibinfo{year}{2015}\natexlab{b}.
\newblock \showarticletitle{ElSe: Ellipse Selection for Robust Pupil Detection in Real-World Environments}.
\newblock \bibinfo{journal}{\emph{Eye Tracking Research and Applications Symposium (ETRA)}}  \bibinfo{volume}{14} (\bibinfo{date}{11} \bibinfo{year}{2015}), \bibinfo{pages}{123--130}.
\newblock
\showISBNx{9781450341257}
\urldef\tempurl%
\url{https://doi.org/10.48550/arxiv.1511.06575}
\showDOI{\tempurl}


\bibitem[Fuhl et~al\mbox{.}(2021)]%
        {Fuhl2021}
\bibfield{author}{\bibinfo{person}{Wolfgang Fuhl}, \bibinfo{person}{Johannes Schneider}, {and} \bibinfo{person}{Enkelejda Kasneci}.} \bibinfo{year}{2021}\natexlab{}.
\newblock \showarticletitle{1000 Pupil Segmentations in a Second using Haar Like Features and Statistical Learning}.
\newblock \bibinfo{journal}{\emph{Proceedings of the IEEE International Conference on Computer Vision}}  \bibinfo{volume}{2021-October} (\bibinfo{date}{2} \bibinfo{year}{2021}), \bibinfo{pages}{3459--3469}.
\newblock
\showISBNx{9781665401913}
\showISSN{15505499}
\urldef\tempurl%
\url{https://doi.org/10.48550/arxiv.2102.01921}
\showDOI{\tempurl}


\bibitem[Fuhl et~al\mbox{.}(2016b)]%
        {Fuhl2016}
\bibfield{author}{\bibinfo{person}{Wolfgang Fuhl}, \bibinfo{person}{Marc Tonsen}, \bibinfo{person}{Andreas Bulling}, {and} \bibinfo{person}{Enkelejda Kasneci}.} \bibinfo{year}{2016}\natexlab{b}.
\newblock \showarticletitle{Pupil detection for head-mounted eye tracking in the wild: an evaluation of the state of the art}.
\newblock \bibinfo{journal}{\emph{Machine Vision and Applications}}  \bibinfo{volume}{27} (\bibinfo{date}{11} \bibinfo{year}{2016}), \bibinfo{pages}{1275--1288}.
\newblock
Issue 8.
\showISSN{14321769}
\urldef\tempurl%
\url{https://doi.org/10.1007/S00138-016-0776-4/FIGURES/14}
\showDOI{\tempurl}


\bibitem[Garbin et~al\mbox{.}(2019)]%
        {Garbin2019}
\bibfield{author}{\bibinfo{person}{Stephan~J. Garbin}, \bibinfo{person}{Yiru Shen}, \bibinfo{person}{Immo Schuetz}, \bibinfo{person}{Robert Cavin}, \bibinfo{person}{Gregory Hughes}, {and} \bibinfo{person}{Sachin~S. Talathi}.} \bibinfo{year}{2019}\natexlab{}.
\newblock \showarticletitle{OpenEDS: Open Eye Dataset}.
\newblock  (\bibinfo{date}{4} \bibinfo{year}{2019}).
\newblock
\urldef\tempurl%
\url{https://arxiv.org/abs/1905.03702v2}
\showURL{%
\tempurl}


\bibitem[GmbH(2022a)]%
        {Pupil2}
\bibfield{author}{\bibinfo{person}{Pupil~Labs GmbH}.} \bibinfo{year}{2022}\natexlab{a}.
\newblock \bibinfo{booktitle}{\emph{HTC Vive Add-On}}.
\newblock
\urldef\tempurl%
\url{https://docs.pupil-labs.com/vr-ar/htc-vive/}
\showURL{%
\tempurl}


\bibitem[GmbH(2022b)]%
        {Pupil}
\bibfield{author}{\bibinfo{person}{Pupil~Labs GmbH}.} \bibinfo{year}{2022}\natexlab{b}.
\newblock \bibinfo{booktitle}{\emph{Pupil core - open source eye tracking platform - pupil labs}}.
\newblock
\urldef\tempurl%
\url{https://pupil-labs.com/products/core/}
\showURL{%
\tempurl}


\bibitem[Guestrin and Eizenman(2006)]%
        {1634506}
\bibfield{author}{\bibinfo{person}{E.D. Guestrin} {and} \bibinfo{person}{M. Eizenman}.} \bibinfo{year}{2006}\natexlab{}.
\newblock \showarticletitle{General theory of remote gaze estimation using the pupil center and corneal reflections}.
\newblock \bibinfo{journal}{\emph{IEEE Transactions on Biomedical Engineering}} \bibinfo{volume}{53}, \bibinfo{number}{6} (\bibinfo{year}{2006}), \bibinfo{pages}{1124--1133}.
\newblock
\urldef\tempurl%
\url{https://doi.org/10.1109/TBME.2005.863952}
\showDOI{\tempurl}


\bibitem[Hennessey et~al\mbox{.}(2006)]%
        {10.1145/1117309.1117349}
\bibfield{author}{\bibinfo{person}{Craig Hennessey}, \bibinfo{person}{Borna Noureddin}, {and} \bibinfo{person}{Peter Lawrence}.} \bibinfo{year}{2006}\natexlab{}.
\newblock \showarticletitle{A single camera eye-gaze tracking system with free head motion}. In \bibinfo{booktitle}{\emph{Proceedings of the 2006 Symposium on Eye Tracking Research \& Applications}} (San Diego, California) \emph{(\bibinfo{series}{ETRA '06})}. \bibinfo{publisher}{Association for Computing Machinery}, \bibinfo{address}{New York, NY, USA}, \bibinfo{pages}{87–94}.
\newblock
\showISBNx{1595933050}
\urldef\tempurl%
\url{https://doi.org/10.1145/1117309.1117349}
\showDOI{\tempurl}


\bibitem[Javadi et~al\mbox{.}(2015)]%
        {Javadi2015}
\bibfield{author}{\bibinfo{person}{Amir~Homayoun Javadi}, \bibinfo{person}{Zahra Hakimi}, \bibinfo{person}{Morteza Barati}, \bibinfo{person}{Vincent Walsh}, {and} \bibinfo{person}{Lili Tcheang}.} \bibinfo{year}{2015}\natexlab{}.
\newblock \showarticletitle{Set: A pupil detection method using sinusoidal approximation}.
\newblock \bibinfo{journal}{\emph{Frontiers in Neuroengineering}}  \bibinfo{volume}{8} (\bibinfo{date}{4} \bibinfo{year}{2015}), \bibinfo{pages}{4}.
\newblock
Issue APR.
\showISSN{16626443}
\urldef\tempurl%
\url{https://doi.org/10.3389/FNENG.2015.00004/ABSTRACT}
\showDOI{\tempurl}


\bibitem[Kar and Corcoran(2017)]%
        {8003267}
\bibfield{author}{\bibinfo{person}{Anuradha Kar} {and} \bibinfo{person}{Peter Corcoran}.} \bibinfo{year}{2017}\natexlab{}.
\newblock \showarticletitle{A Review and Analysis of Eye-Gaze Estimation Systems, Algorithms and Performance Evaluation Methods in Consumer Platforms}.
\newblock \bibinfo{journal}{\emph{IEEE Access}}  \bibinfo{volume}{5} (\bibinfo{year}{2017}), \bibinfo{pages}{16495--16519}.
\newblock
\urldef\tempurl%
\url{https://doi.org/10.1109/ACCESS.2017.2735633}
\showDOI{\tempurl}


\bibitem[Kassner et~al\mbox{.}(2014)]%
        {Kassner2014}
\bibfield{author}{\bibinfo{person}{Moritz Kassner}, \bibinfo{person}{William Patera}, {and} \bibinfo{person}{Andreas Bulling}.} \bibinfo{year}{2014}\natexlab{}.
\newblock \showarticletitle{Pupil: An Open Source Platform for Pervasive Eye Tracking and Mobile Gaze-based Interaction}.
\newblock \bibinfo{journal}{\emph{UbiComp 2014 - Adjunct Proceedings of the 2014 ACM International Joint Conference on Pervasive and Ubiquitous Computing}} (\bibinfo{date}{4} \bibinfo{year}{2014}), \bibinfo{pages}{1151--1160}.
\newblock
\urldef\tempurl%
\url{https://arxiv.org/abs/1405.0006v1}
\showURL{%
\tempurl}


\bibitem[Kim et~al\mbox{.}(2019)]%
        {Kim2019}
\bibfield{author}{\bibinfo{person}{Joohwan Kim}, \bibinfo{person}{Michael Stengel}, \bibinfo{person}{Alexander Majercik}, \bibinfo{person}{Shalini~De Mello}, \bibinfo{person}{David Dunn}, \bibinfo{person}{Samuli Laine}, \bibinfo{person}{Morgan McGuire}, {and} \bibinfo{person}{David Luebke}.} \bibinfo{year}{2019}\natexlab{}.
\newblock \showarticletitle{NVGaze: An anatomically-informed dataset for low-latency, near-eye gaze estimation}.
\newblock \bibinfo{journal}{\emph{Conference on Human Factors in Computing Systems - Proceedings}} (\bibinfo{date}{5} \bibinfo{year}{2019}).
\newblock
\showISBNx{9781450359702}
\urldef\tempurl%
\url{https://doi.org/10.1145/3290605.3300780}
\showDOI{\tempurl}


\bibitem[Kothari et~al\mbox{.}(2022)]%
        {Kothari2022}
\bibfield{author}{\bibinfo{person}{Rakshit~S. Kothari}, \bibinfo{person}{Reynold~J. Bailey}, \bibinfo{person}{Christopher Kanan}, \bibinfo{person}{Jeff~B. Pelz}, {and} \bibinfo{person}{Gabriel~J. Diaz}.} \bibinfo{year}{2022}\natexlab{}.
\newblock \showarticletitle{EllSeg-Gen, towards Domain Generalization for head-mounted eyetracking}.
\newblock \bibinfo{journal}{\emph{Proceedings of the ACM on Human-Computer Interaction}}  \bibinfo{volume}{6} (\bibinfo{date}{5} \bibinfo{year}{2022}).
\newblock
Issue ETRA.
\urldef\tempurl%
\url{https://doi.org/10.1145/3530880}
\showDOI{\tempurl}


\bibitem[Kothari et~al\mbox{.}(2020)]%
        {Kothari2020}
\bibfield{author}{\bibinfo{person}{Rakshit~S. Kothari}, \bibinfo{person}{Aayush~K. Chaudhary}, \bibinfo{person}{Reynold~J. Bailey}, \bibinfo{person}{Jeff~B. Pelz}, {and} \bibinfo{person}{Gabriel~J. Diaz}.} \bibinfo{year}{2020}\natexlab{}.
\newblock \showarticletitle{EllSeg: An Ellipse Segmentation Framework for Robust Gaze Tracking}.
\newblock \bibinfo{journal}{\emph{IEEE Transactions on Visualization and Computer Graphics}}  \bibinfo{volume}{27} (\bibinfo{date}{7} \bibinfo{year}{2020}), \bibinfo{pages}{2757--2767}.
\newblock
Issue 5.
\urldef\tempurl%
\url{https://doi.org/10.1109/tvcg.2021.3067765}
\showDOI{\tempurl}


\bibitem[Li et~al\mbox{.}(2005)]%
        {Li2005}
\bibfield{author}{\bibinfo{person}{Dongheng Li}, \bibinfo{person}{David Winfield}, {and} \bibinfo{person}{Derrick~J. Parkhurst}.} \bibinfo{year}{2005}\natexlab{}.
\newblock \showarticletitle{Starburst: A hybrid algorithm for video-based eye tracking combining feature-based and model-based approaches}.
\newblock \bibinfo{journal}{\emph{IEEE Computer Society Conference on Computer Vision and Pattern Recognition Workshops}}  \bibinfo{volume}{2005-September} (\bibinfo{year}{2005}).
\newblock
\showISBNx{0769526608}
\showISSN{21607516}
\urldef\tempurl%
\url{https://doi.org/10.1109/CVPR.2005.531}
\showDOI{\tempurl}


\bibitem[Ltd(2023)]%
        {SRResearch}
\bibfield{author}{\bibinfo{person}{SR~Research Ltd}.} \bibinfo{year}{2023}\natexlab{}.
\newblock \bibinfo{booktitle}{\emph{SR Research Ltd. - Eye-Tracking Company}}.
\newblock
\urldef\tempurl%
\url{https://www.sr-research.com/}
\showURL{%
\tempurl}


\bibitem[Macinnes et~al\mbox{.}(2018)]%
        {Macinnes299925}
\bibfield{author}{\bibinfo{person}{Jeff~J. Macinnes}, \bibinfo{person}{Shariq Iqbal}, \bibinfo{person}{John Pearson}, {and} \bibinfo{person}{Elizabeth~N. Johnson}.} \bibinfo{year}{2018}\natexlab{}.
\newblock \showarticletitle{Wearable Eye-tracking for Research: Automated dynamic gaze mapping and accuracy/precision comparisons across devices}.
\newblock \bibinfo{journal}{\emph{bioRxiv}} (\bibinfo{year}{2018}).
\newblock
\urldef\tempurl%
\url{https://doi.org/10.1101/299925}
\showDOI{\tempurl}
\showeprint{https://www.biorxiv.org/content/early/2018/06/28/299925.full.pdf}


\bibitem[Mackworth and Thomas(1962)]%
        {mackworth1962head}
\bibfield{author}{\bibinfo{person}{Norman~H Mackworth} {and} \bibinfo{person}{Edward~Llewellyn Thomas}.} \bibinfo{year}{1962}\natexlab{}.
\newblock \showarticletitle{Head-mounted eye-marker camera}.
\newblock \bibinfo{journal}{\emph{JOSA}} \bibinfo{volume}{52}, \bibinfo{number}{6} (\bibinfo{year}{1962}), \bibinfo{pages}{713--716}.
\newblock


\bibitem[Merchant(1967)]%
        {merchant1967oculometer}
\bibfield{author}{\bibinfo{person}{John Merchant}.} \bibinfo{year}{1967}\natexlab{}.
\newblock \bibinfo{booktitle}{\emph{The oculometer}}.
\newblock \bibinfo{type}{{T}echnical {R}eport}.
\newblock


\bibitem[Meyer et~al\mbox{.}(2006)]%
        {10.1007/11768029_25}
\bibfield{author}{\bibinfo{person}{Andr{\'e} Meyer}, \bibinfo{person}{Martin B{\"o}hme}, \bibinfo{person}{Thomas Martinetz}, {and} \bibinfo{person}{Erhardt Barth}.} \bibinfo{year}{2006}\natexlab{}.
\newblock \showarticletitle{A Single-Camera Remote Eye Tracker}. In \bibinfo{booktitle}{\emph{Perception and Interactive Technologies}}, \bibfield{editor}{\bibinfo{person}{Elisabeth Andr{\'e}}, \bibinfo{person}{Laila Dybkj{\ae}r}, \bibinfo{person}{Wolfgang Minker}, \bibinfo{person}{Heiko Neumann}, {and} \bibinfo{person}{Michael Weber}} (Eds.). \bibinfo{publisher}{Springer Berlin Heidelberg}, \bibinfo{address}{Berlin, Heidelberg}, \bibinfo{pages}{208--211}.
\newblock
\showISBNx{978-3-540-34744-6}


\bibitem[Nair(2020)]%
        {Nair2020thesis}
\bibfield{author}{\bibinfo{person}{Nitinraj Nair}.} \bibinfo{year}{2020}\natexlab{}.
\newblock \showarticletitle{RIT-Eyes: Realistic Eye Image and Video Generation for Eye Tracking Applications}.
\newblock \bibinfo{journal}{\emph{Theses}} (\bibinfo{date}{6} \bibinfo{year}{2020}).
\newblock
\urldef\tempurl%
\url{https://scholarworks.rit.edu/theses/10553}
\showURL{%
\tempurl}


\bibitem[Nair et~al\mbox{.}(2020)]%
        {Nair2020}
\bibfield{author}{\bibinfo{person}{Nitinraj Nair}, \bibinfo{person}{Rakshit Kothari}, \bibinfo{person}{Aayush~K. Chaudhary}, \bibinfo{person}{Zhizhuo Yang}, \bibinfo{person}{Gabriel~J. Diaz}, \bibinfo{person}{Jeff~B. Pelz}, {and} \bibinfo{person}{Reynold~J. Bailey}.} \bibinfo{year}{2020}\natexlab{}.
\newblock \showarticletitle{RIT-Eyes: Rendering of near-eye images for eye-tracking applications}.
\newblock \bibinfo{journal}{\emph{Proceedings - SAP 2020: ACM Symposium on Applied Perception}} (\bibinfo{date}{6} \bibinfo{year}{2020}).
\newblock
\urldef\tempurl%
\url{https://doi.org/10.1145/3385955.3407935}
\showDOI{\tempurl}


\bibitem[Santini et~al\mbox{.}(2017)]%
        {Santini2017}
\bibfield{author}{\bibinfo{person}{Thiago Santini}, \bibinfo{person}{Wolfgang Fuhl}, {and} \bibinfo{person}{Enkelejda Kasneci}.} \bibinfo{year}{2017}\natexlab{}.
\newblock \showarticletitle{PuRe: Robust pupil detection for real-time pervasive eye tracking}.
\newblock \bibinfo{journal}{\emph{Computer Vision and Image Understanding}}  \bibinfo{volume}{170} (\bibinfo{date}{12} \bibinfo{year}{2017}), \bibinfo{pages}{40--50}.
\newblock
\urldef\tempurl%
\url{https://doi.org/10.1016/j.cviu.2018.02.002}
\showDOI{\tempurl}


\bibitem[Santini et~al\mbox{.}(2018)]%
        {santini2018purest}
\bibfield{author}{\bibinfo{person}{Thiago Santini}, \bibinfo{person}{Wolfgang Fuhl}, {and} \bibinfo{person}{Enkelejda Kasneci}.} \bibinfo{year}{2018}\natexlab{}.
\newblock \showarticletitle{PuReST: Robust pupil tracking for real-time pervasive eye tracking}. In \bibinfo{booktitle}{\emph{Proceedings of the 2018 ACM symposium on eye tracking research \& applications}}. \bibinfo{pages}{1--5}.
\newblock


\bibitem[Swirski and Dodgson(2013)]%
        {swirski2013fully}
\bibfield{author}{\bibinfo{person}{Lech Swirski} {and} \bibinfo{person}{Neil Dodgson}.} \bibinfo{year}{2013}\natexlab{}.
\newblock \showarticletitle{A fully-automatic, temporal approach to single camera, glint-free 3D eye model fitting}.
\newblock \bibinfo{journal}{\emph{Proc. PETMEI}} (\bibinfo{year}{2013}), \bibinfo{pages}{1--11}.
\newblock


\bibitem[Tonsen et~al\mbox{.}(2015)]%
        {Tonsen2015}
\bibfield{author}{\bibinfo{person}{Marc Tonsen}, \bibinfo{person}{Xucong Zhang}, \bibinfo{person}{Yusuke Sugano}, {and} \bibinfo{person}{Andreas Bulling}.} \bibinfo{year}{2015}\natexlab{}.
\newblock \showarticletitle{Labeled pupils in the wild: A dataset for studying pupil detection in unconstrained environments}.
\newblock \bibinfo{journal}{\emph{Eye Tracking Research and Applications Symposium (ETRA)}}  \bibinfo{volume}{14} (\bibinfo{date}{11} \bibinfo{year}{2015}), \bibinfo{pages}{139--142}.
\newblock
\urldef\tempurl%
\url{https://doi.org/10.1145/2857491.2857520}
\showDOI{\tempurl}


\bibitem[Wang et~al\mbox{.}(2021)]%
        {Wang2021}
\bibfield{author}{\bibinfo{person}{Zhimin Wang}, \bibinfo{person}{Yuxin Zhao}, \bibinfo{person}{Yunfei Liu}, {and} \bibinfo{person}{Feng Lu}.} \bibinfo{year}{2021}\natexlab{}.
\newblock \showarticletitle{Edge-guided near-eye image analysis for head mounted displays}.
\newblock \bibinfo{journal}{\emph{Proceedings - 2021 IEEE International Symposium on Mixed and Augmented Reality, ISMAR 2021}} (\bibinfo{year}{2021}), \bibinfo{pages}{11--20}.
\newblock
\showISBNx{9781665401586}
\urldef\tempurl%
\url{https://doi.org/10.1109/ISMAR52148.2021.00015}
\showDOI{\tempurl}


\bibitem[Wood et~al\mbox{.}(2016)]%
        {wood2016_etra}
\bibfield{author}{\bibinfo{person}{Erroll Wood}, \bibinfo{person}{Tadas Baltru{\v{s}}aitis}, \bibinfo{person}{Louis-Philippe Morency}, \bibinfo{person}{Peter Robinson}, {and} \bibinfo{person}{Andreas Bulling}.} \bibinfo{year}{2016}\natexlab{}.
\newblock \showarticletitle{Learning an Appearance-Based Gaze Estimator from One Million Synthesised Images}. In \bibinfo{booktitle}{\emph{Proceedings of the Ninth Biennial ACM Symposium on Eye Tracking Research \& Applications}}. \bibinfo{pages}{131--138}.
\newblock


\bibitem[Yiu et~al\mbox{.}(2019)]%
        {Yiu2019}
\bibfield{author}{\bibinfo{person}{Yuk~Hoi Yiu}, \bibinfo{person}{Moustafa Aboulatta}, \bibinfo{person}{Theresa Raiser}, \bibinfo{person}{Leoni Ophey}, \bibinfo{person}{Virginia~L. Flanagin}, \bibinfo{person}{Peter zu Eulenburg}, {and} \bibinfo{person}{Seyed~Ahmad Ahmadi}.} \bibinfo{year}{2019}\natexlab{}.
\newblock \showarticletitle{DeepVOG: Open-source pupil segmentation and gaze estimation in neuroscience using deep learning}.
\newblock \bibinfo{journal}{\emph{Journal of Neuroscience Methods}}  \bibinfo{volume}{324} (\bibinfo{date}{8} \bibinfo{year}{2019}).
\newblock
\showISSN{1872678X}
\urldef\tempurl%
\url{https://doi.org/10.1016/J.JNEUMETH.2019.05.016}
\showDOI{\tempurl}


\bibitem[Zhu and Ji(2005)]%
        {ZHU2005124}
\bibfield{author}{\bibinfo{person}{Zhiwei Zhu} {and} \bibinfo{person}{Qiang Ji}.} \bibinfo{year}{2005}\natexlab{}.
\newblock \showarticletitle{Robust real-time eye detection and tracking under variable lighting conditions and various face orientations}.
\newblock \bibinfo{journal}{\emph{Computer Vision and Image Understanding}} \bibinfo{volume}{98}, \bibinfo{number}{1} (\bibinfo{year}{2005}), \bibinfo{pages}{124--154}.
\newblock
\showISSN{1077-3142}
\urldef\tempurl%
\url{https://doi.org/10.1016/j.cviu.2004.07.012}
\showDOI{\tempurl}
\newblock
\shownote{Special Issue on Eye Detection and Tracking}.


\bibitem[Świrski et~al\mbox{.}(2012)]%
        {Swirski2012}
\bibfield{author}{\bibinfo{person}{Lech Świrski}, \bibinfo{person}{Andreas Bulling}, {and} \bibinfo{person}{Neil Dodgson}.} \bibinfo{year}{2012}\natexlab{}.
\newblock \showarticletitle{Robust real-time pupil tracking in highly off-axis images}.
\newblock \bibinfo{journal}{\emph{Eye Tracking Research and Applications Symposium (ETRA)}} (\bibinfo{year}{2012}), \bibinfo{pages}{173--176}.
\newblock
\showISBNx{9781450312257}
\urldef\tempurl%
\url{https://doi.org/10.1145/2168556.2168585}
\showDOI{\tempurl}


\end{thebibliography}

\end{document}